\documentclass[journal]{IEEEtran}
\usepackage{color}
\usepackage{xcolor}

\usepackage{xr}
\makeatletter
\newcommand*{\addFileDependency}[1]{
  \typeout{(#1)}
  \@addtofilelist{#1}
  \IfFileExists{#1}{}{\typeout{No file #1.}}
}
\makeatother

\newcommand*{\myexternaldocument}[1]{%
    \externaldocument{#1}%
    \addFileDependency{#1.tex}%
    \addFileDependency{#1.aux}%
}

\myexternaldocument{tmi_supplementary}

\usepackage{tmi}

\usepackage{amsmath,amssymb,amsfonts}
\usepackage{algorithmic}
\usepackage{graphicx}
\usepackage{xspace}
\usepackage{subcaption}
\usepackage{bbold}
\usepackage{textcomp}
\usepackage{mathtools}

\usepackage[numbers, compress]{natbib}

\usepackage{amsmath,amsfonts,bm}

\def\eqref#1{equation~\ref{#1}}

\def\1{\bm{1}}

\def\vs{{\bm{s}}}

\DeclareMathAlphabet{\mathsfit}{\encodingdefault}{\sfdefault}{m}{sl}
\SetMathAlphabet{\mathsfit}{bold}{\encodingdefault}{\sfdefault}{bx}{n}

\def\gI{{\mathcal{I}}}

\def\gL{{\mathcal{L}}}

\def\gP{{\mathcal{P}}}
\def\gQ{{\mathcal{Q}}}

\def\BibTeX{{\rm B\kern-.05em{\sc i\kern-.025em b}\kern-.08em
    T\kern-.1667em\lower.7ex\hbox{E}\kern-.125emX}}
\makeatletter
\DeclareRobustCommand\onedot{\futurelet\@let@token\@onedot}
\def\@onedot{\ifx\@let@token.\else.\null\fi\xspace}

\def\ie{\emph{i.e}\onedot}

\def\vs{\emph{vs}\onedot}

\makeatother

\colorlet{CLRBlue}{black}

\begin{document}

\title{Adaptive Contrast for Image Regression in Computer-Aided Disease Assessment}
\author{Weihang Dai, Xiaomeng Li, \IEEEmembership{Member, IEEE}, Wan Hang Keith Chiu,\\ Michael D. Kuo and Kwang-Ting Cheng, \IEEEmembership{Fellow, IEEE}

\thanks{Manuscript received XX XX, 2020. 

W. Dai is with the Department of Computer Science and Engineering, The Hong Kong University of Science and Technology, Hong Kong SAR, China (e-mail:wdai03@gmail.com). 

W.H.K Chiu and M.D. Kuo are with the Department of Diagnostic Radiology, LKS Faculty of Medicine, The Hong Kong University, Hong Kong SAR, China (e-mails: kwhchiu@hku.hk, mikedkuo@gmail.com).

X. Li is with the Department of Electronic and Computer Engineering, The Hong Kong University of Science and Technology, Hong Kong SAR, China, and also with The Hong Kong University of Science and Technology Shenzhen Research Institute (e-mails: eexmli@ust.hk)
(Corresponding author: Xiaomeng Li.)

K.T. Cheng is with the Department of Computer Science and Engineering and the Department of Electronic and Computer Engineering, The Hong Kong University of Science and Technology, Hong Kong SAR, China (e-mail: timcheng@ust.hk). 
} 
\thanks{Copyright (c) 2021 IEEE. Personal use of this material is permitted. Permission from IEEE must be obtained for all other uses, including reprinting/republishing this material for advertising or promotional purposes, collecting new collected works for resale or redistribution to servers or lists, or reuse of any copyrighted component of this work in other works.}
}

\maketitle

\newcommand{\xmli}[1]{{\color{blue}{[XM: #1]}}}
\newcommand{\wdai}[1]{{\color{red}{[WD: #1]}}}
\newcommand{\wdairv}[1]{{\color{CLRBlue}{#1}}}

\begin{abstract}

Image regression tasks for medical applications, such as bone mineral density (BMD) estimation and left-ventricular ejection fraction (LVEF) prediction, play an important role in computer-aided disease assessment. Most deep regression methods train the neural network with a single regression loss function like MSE or L1 loss. 
In this paper, we propose the first contrastive learning framework for deep image regression, namely AdaCon, which consists of a feature learning branch via a novel adaptive-margin contrastive loss and a regression prediction branch.
Our method incorporates label distance relationships as part of the learned feature representations, which allows for better performance in downstream regression tasks. Moreover, it can be used as a plug-and-play module to improve performance of existing regression methods. 
We demonstrate the effectiveness of AdaCon on two medical image regression tasks, \ie, bone mineral density estimation from X-ray images and left-ventricular ejection fraction prediction from echocardiogram videos. AdaCon leads to relative improvements of 3.3\% and 5.9\% in MAE over state-of-the-art BMD estimation and LVEF prediction methods, respectively. 

\end{abstract}


\section{Introduction}
\label{sec:introduction}

Medical image classification and segmentation with convolutional neural networks (CNNs) have seen huge adoption in various applications for computer-aided diagnosis, such as disease grading~\cite{li2019canet, xie2021cross}, organ segmentation~\cite{huang2020unet, gibson2018automatic}, and tumor segmentation~\cite{kamnitsas2015multi, kamnitsas2017efficient,li2018h,li2020transformation}. Comparatively, medical image regression tasks based on CNNs have received less attention.
Unlike classification and segmentation problems, regression tasks generate continuous value predictions that are important indicators for various disease assessments in clinical practice.
Two examples of this are bone mineral density (BMD) and left-ventricular ejection fraction (LVEF). BMD is used to identify osteoporosis, determine the risk of fractures, and measure patient response to osteoporosis treatment~\cite{kranioti2019bone}. 
Typical BMD values for the Asian population range between 0.5-0.9, with a low reading indicating loss of bone mass and higher risks of fractures. 
LVEF is a percentage measurement of how much blood the left ventricle pumps out with each contraction, serving as an essential indicator to diagnose and track heart failure. 
Normal LVEF ranges from 55\% to 70\%, with LVEF higher than 75\% or lower than 40\% indicating a potential for heart failure or cardiomyopathy~\cite{maeder2009heart}. 
Early detection and accurate assessment of BMD and LVEF are important for early intervention in clinical practice.

Early works for image regression relied on hand-crafted filters as feature inputs in a multivariate regression model. For example,  Pulkkinen \textit{et al.}~\cite{pulkkinen2008experimental} estimate BMD by applying a gradient filter to patches of femur-neck X-ray images and calculating summary statistics as feature inputs. 
Similar techniques have also been applied for computer vision tasks using natural images, such as facial age estimation~\cite{el2011human, kim2015facial} and single-image depth estimation~\cite{saxena2005learning}. 
Recent progress in image regression has increasingly featured deep learning techniques~\cite{luo2017multi,ouyang2020video,zheng2021semi}. For instance, Luo \textit{et al.}~\cite{luo2017multi} used a VGG backbone~\cite{simonyan2014very} on cardiac magnetic resonance volumes to directly regress ventricular volumes. Ouyang \textit{et al.}~\cite{ouyang2020video} employed the R2+1D ResNet network~\cite{tran2018closer} to directly regress LVEF values from echocardiogram videos. 
Among these works, mean squared error (MSE) is used as the loss function to measure prediction error and supervise training.
Alternative loss functions such as L1 loss~\cite{hsieh2021automated}, Huber Loss~\cite{huber1992robust, carvalho2018regression}, and Tukey loss~\cite{belagiannis2015robust} 
have also been proposed for image regression as ways to reduce the effect of sample outliers.

Although different regression loss functions have been explored, model performance from training with a singular \wdairv{regression loss function can still be limited}. Recently, contrastive learning has shown great promise for feature representation learning in a variety of computer vision tasks, such as unsupervised pretraining~\cite{chen2020simple,caron2020unsupervised,li2020prototypical,chen2020improved}, image classification~\cite{khosla2020supervised}, and video detection~\cite{dave2021tclr}. For example, Chen \textit{et al.}~\cite{chen2020simple} showed that their unsupervised SimCLR framework achieves comparable performance with state-of-the-art performance on ImageNet classification.
Khosla \textit{et al.}~\cite{khosla2020supervised} further demonstrated that contrastive learning used in a supervised setting could surpass state-of-the-art ImageNet results.
Although contrastive learning has demonstrated excellent performance for various tasks, adapting it for image regression problems still remains under-explored.

In this paper, our goal is to explore the feasibility of contrastive learning for image regression tasks. 
One na\"ive solution is to treat each continuous value as an individual class and use the supervised contrastive loss function to learn feature representations~\cite{khosla2020supervised}. 
However, this method ignores the underlying distance relationships between labels, leading to inferior results; see results of SupCon in Tables \ref{loss_comp} and \ref{loss_comp2}.
To this end, we present a simple and effective framework, namely AdaCon, that introduces a novel distance-adaptive contrastive loss function to model distance relationships in the representation space for image regression tasks. AdaCon conducts contrastive learning and regression prediction in a multi-task framework and learns features that are more consistent with the regression task, thus improving the overall prediction performance.
We demonstrate our methodology on two medical image regression tasks, \ie, BMD estimation from X-rays images and LVEF prediction from echocardiogram videos. 

The key contributions of this study are: 

\begin{itemize}
    \item We present a simple yet effective framework (AdaCon) for image regression tasks, consisting of a contrastive feature learning branch for representation learning and a regression estimation branch for regression. Our method can serve as a plug-and-play component in existing methods to improve image regression performance. 
    
    \item AdaCon introduces a novel adaptive-margin loss function to preserve distance and order relationships among continuous labels in the representation space. We show that the enhanced feature representation can improve image regression results. 
    
    \item AdaCon achieves state-of-the-art performance on two medical image regression tasks: BMD estimation from single X-ray images and LVEF prediction from echocardiogram videos\wdairv{\footnote{\wdairv{Code is available at https://github.com/XMed-Lab/AdaCon}}}. Ablation studies demonstrate AdaCon's superior performance over other methods\wdairv{. Results are also validated using three additional datasets. }
\end{itemize}

\section{Related Work}

In this section, we review contrastive learning for representation learning and some of its applications. 
We further discuss related works for BMD estimation and LVEF prediction. 

\subsection{Contrastive Learning}

\noindent \textbf{Unsupervised representation learning.}
In the early stages of unsupervised representation learning, researchers relied on dimension-reduction techniques such as PCA~\cite{bengio2013representation}, and clustering-based methods on image patches~\cite{singh2012unsupervised, doersch2013mid, huang2016unsupervised} to extract feature representations.
With the popularity of deep learning techniques, architectures like deep auto-encoders, which introduce dimension reducing bottlenecks and are trained by minimizing reconstruction error~\cite{hinton2006fast, hinton2006reducing, kingma2013auto}, have become increasingly used. 
Studies have also explored pretraining networks on pretext tasks such as image transformation identification, showing that it can be an effective way to learn feature representations for images~\cite{noroozi2016unsupervised, gidaris2018unsupervised, zhang2019aet}. The pretrained networks can then be applied to downstream tasks by training a classifier on top of the feature extractor. 

In recent years, however, contrastive learning methods have dominated this field. Chen \textit{et al.}~\cite{chen2020simple} proposed the SimCLR framework, in which a network is pretrained to identify positive sample pairs, defined as samples augmented from the same source image. They demonstrated that when fine-tuning the pretrained network to perform downstream image classification tasks, they can achieve state-of-the-art performance, sometimes even exceeding results from supervised learning. Similarly, MoCo identifies augmented images from the same source by maximizing feature similarity between positive pairs using a momentum encoder, thus reducing the reliance on large batch sizes \cite{he2020momentum}. Subsequent variations adopt improvements from SimCLR and vision transformers into Moco\_v2~\cite{chen2020improved} and Moco\_v3~\cite{chen2021empirical}. Additional studies have also looked at improving performance through stronger augmentations~\cite{tian2020makes}, adapting underlying concepts to different architectures~\cite{you2020graph}, or using contrastive learning for unsupervised tasks like clustering ~\cite{caron2020unsupervised}.

\noindent \textbf{Contrastive learning for downstream applications.}
Given the stunning improvements seen in unsupervised representation learning, researchers started to adapt contrastive methods to various downstream tasks such as image classification~\cite{khosla2020supervised, wang2021contrastive}, segmentation~\cite{wang2021dense, zhao2020contrastive}, and video detection~\cite{dave2021tclr}. Khosla \textit{et al.}~\cite{khosla2020supervised} showed that by using label information to identify additional positive-pair samples through supervised contrastive learning, classification accuracy on the benchmark ImageNet classification can be improved by 0.8\%. Wang \textit{et al.}~\cite{wang2021contrastive} also demonstrated that contrastive learning can be used to improve classification performance for highly imbalanced datasets. Wang and Zhao \textit{et al.}~\cite{wang2021dense, zhao2020contrastive} demonstrated that contrastive learning can be used to improve performance on image segmentation tasks by performing contrastive learning on pixel-level features instead of global features. By considering temporal information, Dave \textit{et al.}~\cite{dave2021tclr} also showed that contrastive learning can also be used efficiently to learn feature representations for videos. 

Although contrastive learning has been applied to various tasks, adapting it to regression remains an open and non-trivial problem. This study proposes a simple adaptive contrastive learning framework, AdaCon, that allows the same principles to be applied to continuous value prediction. We demonstrate that AdaCon can be used as a general technique to improve performance for different image regression tasks.

\noindent \textbf{Comparison with deep metric learning.}
Contrastive learning is closely related to traditional \wdairv{distance metric learning}, which uses margins to impose separation between different classes in the feature space~\cite{chopra2005learning, weinberger2009distance, schroff2015facenet, sohn2016improved}. The main difference between the two is that contrastive learning generalizes to an arbitrary number of positive and negative sample pairs. In contrast, metric learning loss functions, such as triplet loss~\cite{weinberger2009distance} and N-pair loss~\cite{sohn2016improved}, are restricted in the number of pairs they use. Triplet loss, in particular, is known to be unstable and highly dependent on hard-sample mining~\cite{schroff2015facenet,sohn2016improved,yu2018correcting}. The number of possible triplet combinations also scales at a cubic rate relative to sample size, making it difficult to sample well for large batch sizes. 
Khosla \textit{et al.}~\cite{khosla2020supervised} show analytically that the SupCon loss function is a more general form of metric learning loss that reduces to triplet loss and N-pair loss under certain conditions and is a superior alternative. They also show empirically that SupCon outperforms these losses due to the inherent hard-mining properties of its loss function. 

\wdairv{Distance metric learning} has been successfully used in regression problems by making the separation margins adaptive. The adaptive triplet loss, for example, imposes larger margins on samples pairs with larger differences in values~\cite{zheng2021semi,zhao2019weakly}.
However, these functions still rely on triplet sampling and only applies constraints within each sampled triplet group. 
Unlike this method, the optimization of our method is performed over an entire batch with constraints applied simultaneously across all samples, contributing to a more stable training process and better feature representations. We also empirically show in Tables \ref{loss_comp} and \ref{loss_comp2} that AdaCon outperforms adaptive triplet loss.

\subsection{Bone Mineral Density Estimation from Plain X-ray Films}

Medical imaging studies on bone X-rays have traditionally been focused on fracture detection~\cite{zhang2021new, krogue2020automatic, guan2020arm}, disease diagnosis~\cite{huang2020radiographic, lu2018deep}, or segmentation~\cite{lu2018deep,sekuboyina2017attention}. Little attention has been given to bone mineral density (BMD) estimation, which is important for diagnosing osteoporosis, a condition characterized by decreased levels of bone \wdairv{density}. BMD is calculated as bone-mass/coverage-area ($g/cm^2$) over an region of interest, typically the femur-neck or spine, and the gold standard is measurement by dual-energy X-ray absorptiometry (DXA). However, due to the limited availability of DXA devices, especially in developing countries, osteoporosis is often under-diagnosed and under-treated. 
Compared to DXA, plain X-ray films are easier and cheaper to obtain.  Therefore, alternative lower-cost BMD evaluation methods using more accessible mediums like X-ray films are highly beneficial.

Early works explored using handcrafted features from X-rays to correlate BMD. Pulkkinen and Chappard ~\cite{pulkkinen2008experimental, chappard2010prediction} for example both calculate texture-based summary statistics on X-ray patches to regress on BMD using multi-variate regression. Deep learning methodologies have also been used recently following their popularity in computer vision problems. Chu \textit{et al.}~\cite{chu2018using} use a siamese network on patches from dental radiographs to perform osteoporosis diagnosis. Wang \textit{et al.}~\cite{wang2021opportunistic} use a GCN on chest X-rays to first perform ROI detection before regressing image crops on BMD with a VGG-16 network and MSE loss. Hsieh \textit{et al.}~\cite{hsieh2021automated} use a VGG-11 network and L1 loss to regress BMD from femur-neck and spine vertebrate crops after performing ROI detection. Zheng \textit{et al.}~\cite{zheng2021semi} use a VGG architecture for BMD regression with additional adaptive triplet loss and semi-supervised learning.

Most of the existing techniques for BMD estimation only use a single regression loss to supervise the training process with no further optimization considered on the feature space. In contrast to these methodologies, we present a novel plug-and-play framework, AdaCon, that can be used for different regression tasks. By performing BMD regression and feature representation learning with AdaCon in a multi-task approach, we can outperform alternative methods \wdairv{by ensuring better features are learnt}, as seen in Table \ref{loss_comp}. Notably, Zheng \textit{et al.}~\cite{zheng2021semi} use the adaptive triplet loss function in addition to regression loss for BMD estimation. We show analytically and empirically that contrastive learning with AdaCon is a better alternative to their methodology.

\subsection{Ejection Fraction Prediction from Echocardiogram Videos}

Ejection fraction is the percentage change of a patient's left-ventricular volume between the end-diastole phase (EDV) and the end-systole phase (ESV). It is an indicator for the heart's blood-pumping capability and is commonly used to diagnose heart failure~\cite{loehr2008heart}. Measurement is normally done manually by outlining the left ventricle in echocardiograms during different stages of a heart beat and estimating the corresponding volume.  
This process is laborious and can also lead to significant inter-observer variation~\cite{ouyang2020video, pellikka2018variability}, therefore driving the need for automated techniques. 

Automated LVEF prediction has been studied using both still imaging and video inputs. The most common approach is to first estimate EDV and ESV volumes from still frames and then calculate LVEF based on these values. Luo \textit{et al.}~\cite{luo2017multi} use cardiac magnetic resonance data to perform deep regression with MSE loss for volume estimation after image pre-processing and slice selection. Zhen \textit{et al.}~\cite{zhen2014direct} use a regression forest model to jointly regress the left and right ventricle volumes from MRI images with MSE. An alternative technique is to treat volume estimation as a segmentation problem and use the segmented area to estimate volume. Jafari \textit{et al.}~\cite{jafari2021deep} use Bayesian estimation for ventricle segmentation from ultra-sound echo cines for volume estimation. Liu \textit{et al.}~\cite{liu2021deep} introduce a pyramid local attention module to capture local feature similarities to improve segmentation accuracy. 

Instead of learning from still images, Ouyang \textit{et al.}~\cite{ouyang2020video} proposed to predict LVEF values directly from echocardiogram videos using their collected dataset, EchoNet-Dynamic. 
They regress video inputs directly using a R2+1D ResNet~\cite{tran2018closer} backbone against LVEF values with MSE loss. Reynaud \textit{et al.}~\cite{reynaud2021ultrasound} also explored LVEF prediction on the EchoNet-Dynamic dataset by using a video transformer network to identify ED and ES frames, which is then used to perform segmentation and volume estimation.

Current deep regression methodologies employ the MSE loss to train the network and do not consider optimizations on the feature space. Unlike these methodologies, we design a new framework, AdaCon, that \wdairv{improves image} regression prediction via \wdairv{an improved} feature representation learning.

\section{Methodology}

We denote $\mathcal{D} \wdairv{\coloneqq} \left \{ (x_i, y_i) \right \}_{i=1}^{N} $ as the training dataset consisting of $N$ image-label pairs, where $y_i$ is the label of image $x_i$.
Our goal for contrastive learning is to learn a feature embedding network $f_{\theta}(\cdot)$ from the training dataset. The network $f_{\theta}(\cdot)$ maps the input $x_i$ to a \wdairv{L2-normalized $d$-dimensional embedding, $z_i$, such that $z_i =  f_{\theta}(x_i) \in \mathbb {S}^{d-1}$ and lies on the unit hypersphere}. 
Let $\gI$ denote the sample indices for a randomly sampled batch during training.
We denote $\gP(i) \wdairv{\coloneqq} \{j\in\gI\:|\:y_j=y_i,\: j\neq i\}$ as the set of indices for all positive samples for anchor sample $i$, \ie, samples with the same label as $y_i$. 
The negative pairs in the batch are images with different labels to anchor sample $i$, the indices for which can be denoted as:
$\gQ(i) \wdairv{\coloneqq} \{j\in\gI\:|\:y_j \neq y_i\}$.

One na\"ive approach to contrastive learning for regression is to apply image augmentations to a batch, treat each continuous label as an individual class, and apply SupCon loss~\cite{khosla2020supervised} directly. 
Then, the loss function can be expressed as:

\begin{equation}\label{eqn:supcon}
   \mathcal{L}_{sup} = \sum_{i\in \gI} \frac{-1}{|\gP(i)|} \sum_{p \in \gP(i)} {\rm log} \frac{ {\rm exp}(s \: \wdairv{\cos}(\theta_{i,p}))}{\sum_{a \in {\gI \setminus i}} {\rm exp}( s \: \wdairv{\cos}(\theta_{i,a}))} \wdairv{\:,}
\end{equation}
where $s$ denotes the temperature scaling factor, usually determined empirically, and $\wdairv{\cos}(\theta_{i,j}) = z_i^T z_j$.

Intuitively, for an anchor sample with label 0.5 and negative samples of labels 0.4 and 0.1, the sample with label 0.4 should be closer to the anchor image than the one with label 0.1 in the representation space. 
However, by using the SupCon loss function directly, \textit{both negative samples are treated the same} during optimization, leading to inferior results.
We can see this empirically in Tables \ref{loss_comp} and \ref{loss_comp2}.

To tackle this problem, we must also capture the relationship between the continuous labels in the representation space. For example, given a randomly sampled batch $\mathcal{I}$ with anchor sample $i$ and samples $j,k$ such that $y_i \leq y_j \leq y_k$, the expected relationship between the sample feature projections is:

\begin{equation}\label{eqn:cosrelate}
        \wdairv{\cos}(\theta_{i,j}) \geq \wdairv{\cos}(\theta_{i,k}) \wdairv{\:.}
\end{equation}

This is because we expect the sample with label closer to the anchor $i$ to have features that are more similar. Equation (\ref{eqn:cosrelate}) should also be true for samples where $ y_i \geq y_j \geq y_k $. We can apply adaptive margins at the decision boundaries between different sample pairs to encourage our model to learn this relationship. By incorporating label distance and ranking information into the learned representations, the features learned will be more consistent with the regression task.

\subsection{Adaptive Contrast Method}
\label{adacon_method}

\begin{figure*}%
\centering
\begin{subfigure}{\columnwidth}
\centering
\includegraphics[width=0.85 \columnwidth]{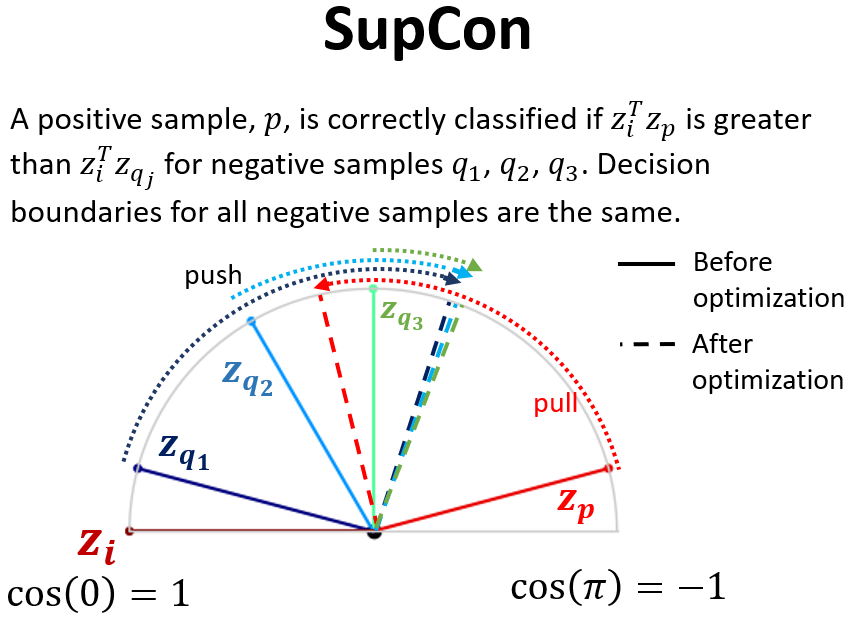}%
\caption{}%
\label{log_subfiga}%
\end{subfigure}\hfill%
\begin{subfigure}{\columnwidth}
\centering
\includegraphics[width=0.85 \columnwidth]{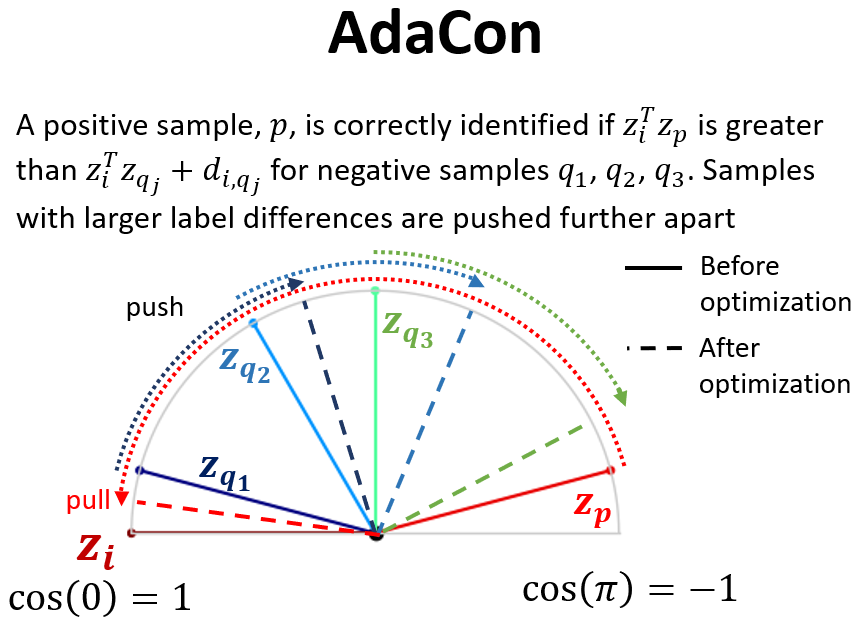}%
\caption{}%
\label{log_subfigb}%
\end{subfigure}\hfill%

\caption{\wdairv{(a)} For classification based supervised contrastive loss, SupCon~\cite{khosla2020supervised}, samples that form positive pairs are pulled closer together in the angular space, whilst negative pairs are pushed further apart. The boundary condition for correct positive-pair identification is $z_i^T z_p > z_i^T z_{q_j}$, \ie  $\wdairv{\cos}(\theta_{i,p}) > \wdairv{\cos}(\theta_{i,q_j})$, where $p \in \gP$ and $q_j \in \gQ$; negative pairs are pushed apart indiscriminately. (b) By adding adaptive margins to the boundary conditions for classification, we enforce the model such that the feature projections $z_{q_2}$ is pushed further apart on the angular space compared to $z_{q_1}$ if $y_p<y_{q_1} < y_{q_2}$ or $y_{q_2}>y_{q_1} > y_{p}$.}
\label{log_graph}
\end{figure*}

In contrastive learning for supervised settings, we identify positive sample pairs, defined as pairs with the same labels, amongst a randomly sampled batch. We add an adaptive margin $d_{i,\wdairv{q}}$ into our decision boundary to introduce label ordering and distance information. For any positive sample $p \in \gP(i)$ and negative sample $q \in \gQ(i)$, we would like to achieve the following:

\begin{equation}\label{eqn:adaconstrain}
    \wdairv{\cos}(\theta_{i,p}) > \wdairv{\cos}(\theta_{i,q}) + d_{i,\wdairv{q}} \wdairv{\:.}
\end{equation}
We want to choose our adaptive margin function, $d_{i,q}$, based on the following characteristics: 

\begin{enumerate}
    
    \item For any $i,j,k \in \gI$ where $y_i < y_j < y_k$, or $y_i > y_j > y_k$, we have $d_{i,k} > d_{i,j}$. This means we want a larger margin imposed on the feature similarity of sample pair ($i,k$) compared to ($i,j$) because sample $k$ is further away from $i$ in label space. 
    
    \item Because the cosine function is bounded between $(-1,1)$, we want $d_{i,\wdairv{q}}$ to be dispersed within the range of $(0,2)$ to encourage feature separation for stronger representation learning.

\end{enumerate}
Based on these characteristics, we define the adaptive margin $d_{i,\wdairv{q}}$ as follows: 

\begin{equation}\label{eqn:adamargin}
        d_{i,\wdairv{q}} = 2 \times | \phi(y_i) - \phi (y_\wdairv{q}) | \wdairv{\:.}
\end{equation}
where $\phi$ is the empirical cumulative distribution function (ECDF)~\cite{van2000asymptotic}, defined by:

\begin{equation}\label{eqn:ecdf}
    \phi (y_i) = \frac{1}{N} \sum_{j=1}^N \mathbb{1}_{y_j\leq y_i} \wdairv{\:.}
\end{equation}

It is easy to see that our adaptive margin function $d_{i,\wdairv{q}}$ satisfies both conditions above. \wdairv{Condition one is satisfied because} $\phi$ is a monotonically increasing function and $\phi (y_i) < \phi (y_j) < \phi (y_k)$ \wdairv{if} $y_i < y_j < y_k$. \wdairv{Therefore,} margin $d_{i,k}$ is greater than $d_{i,j}$. This is also true \wdairv{if} $y_i > y_j > y_k$\wdairv{, because we have $\phi (y_i) > \phi (y_j) > \phi (y_k)$. The }ECDF also provides a natural mapping of label values onto the (0,1) range by normalizing the label distribution to a uniform distribution \wdairv{\cite{10.2307/2132726}}. This ensures margin values are robust to different label distributions, even if they are heavily skewed.
\wdairv{Condition two is satisfied because the} range of $d_{i,\wdairv{q}}$ is within (0,2), which encourages feature separation.

\wdairv{We can incorporate our adaptive margins into the supervised contrastive loss function in (\ref{eqn:supcon}) by adding it after the cosine similarity term, giving us: }

\begin{equation}\label{eqn:adacon_loss}
     \mathcal{L}_{con} = \sum_{i\in \gI} \frac{-1}{|\gP (i)|} \sum_{p \in \gP (i)} {\rm log} \frac{{\rm exp}(s \: \wdairv{\cos(\theta_{i,p})} )}{\sum_{a \in \gI \setminus i} {\rm exp}(s(\wdairv{\cos}(\theta_{i,a}) + d_{i,a}))} \wdairv{,}
\end{equation}
\wdairv{where $d_{i,a}$ follows definition (\ref{eqn:adamargin}). The loss function is minimized when all positive pairs are correctly identified within sample batch $\mathcal{I}$, \ie{} when condition (\ref{eqn:adaconstrain}) is satisfied, thus enforcing our adaptive decision boundary constraints. A more detailed derivation of (\ref{eqn:adacon_loss}) is provided in Appendix \ref{sec:ada_derive}. }
We use this adaptive-margin loss function to supervise contrastive feature learning in our AdaCon framework. Fig.~\ref{log_graph} illustrates AdaCon's loss function as compared to SupCon using \wdairv{a simplified} example on the 2D plane.

\textbf{Margin analysis.} The ECDF of some value $y$, $\phi(y)$, is the sample estimate of $Pr(Y<y)$. \wdairv{Using this as part of our adaptive margin function gives us a number of desirable properties.} 

The first \wdairv{property} is that \wdairv{the} ECDF reduces the effect of sample outliers on margin values because of distribution normalization \wdairv{\cite{10.2307/2132726}}. For example, for some outlying value $y_j$, its relative order ranking will be preserved by the adaptive margin $d_{i,j}$ after transformation, but the margin will not be excessively large because of large differences in \wdairv{label value}. \wdairv{This allows for greater feature separation within the $(-1,1)$ range for remaining samples}. 

The second \wdairv{property} is that \wdairv{the ECDF transformation gives an intuitive mapping of the adaptive margin function based on sample probability}. For example, for samples $i,j$ where $y_i > y_j$, we can substitute \wdairv{$d_{i,j}$ with $2 \times | \phi (y_i) -  \phi (y_j) |$ at the decision boundary to give us}:
\begin{equation}
\wdairv{\cos}(\theta_{i,p}) - \wdairv{\cos}(\theta_{i,j}) = 2 \times | \phi (y_i) -  \phi (y_j) | \wdairv{\:.}
\end{equation}
The value $|\phi (y_i) - \phi(y_j)|$ \wdairv{is equal to} the sample estimate of $Pr(y_j < Y < y_i)$. \wdairv{Therefore, at the decision boundary, the margin between the cosine similarity values 
is proportional to the sample probability of observing a value between $y_j$ and $y_i$. This means if there are a larger number of observed samples with labels between $y_j$ and $y_i$, we will impose a larger margin between $\wdairv{\cos}(\theta_{i,p})$ and $\wdairv{\cos}(\theta_{i,j})$. If there are fewer samples, the margin will be smaller. }

\subsection{Comparison with Adaptive Triplet Loss}

Adaptive methods for \wdairv{distance metric learning} have been used in~\cite{zheng2021semi, zhao2019weakly} for regression tasks by introducing variable separation margins based on label differences into the standard triplet loss function. The major shortfall of triplet loss however is that it is highly dependant on the triplets sampled, but the number of possible triplets grows at a cubic rate relative to batch size, making it difficult \wdairv{to do so effectively}~\cite{schroff2015facenet, sohn2016improved, yu2018correcting}. Margins are also calculated and applied only within each triplet group. Unlike adaptive triplet loss, AdaCon simultaneously applies constraints to all samples within a batch, optimizing their features simultaneously during the training process. Also, the contrastive task identifies positive pairs and therefore only grows at a quadratic rate relative to batch size. It is generalizable to multiple positive and negative pairs. 

We analytically demonstrate in \wdairv{Appendix \ref{sec:ada_1pair}} that AdaCon can be regarded as an improved, general form of adaptive \wdairv{distance} metric learning by showing that the adaptive-margin loss function reduces to an approximation of adaptive triplet loss in the case of only \wdairv{one} positive and \wdairv{one} negative pair. We also support this empirically by demonstrating in our experiments that AdaCon outperforms adaptive triplet loss, as can be seen in Tables \ref{loss_comp} and \ref{loss_comp2}.

\subsection{Full Framework}

The network architecture in our proposed framework consists of a feature extraction backbone and two prediction branches: one for regression and one for contrastive learning. 

The contrastive learning branch outputs feature projections and is composed of a dense layer, a ReLu activation layer, a final dense layer with an output dimension of 128, and a L2 normalization operation. For each input batch $\{x_i; i\in \gI \}$, input samples undergo augmentation to form an augmented batch $\{x'_i; i \in \gI\}$. The two batches are used together in the AdaCon loss function to calculate contrastive loss, $ \mathcal{L}_{con}$. Positive pairs are defined as inputs with the same ground truth labels, which in most cases will be augmented images from the same sample source, \ie, $\{x_i, x'_i\}$. 

\begin{figure}
\centering

\includegraphics[width=1 \columnwidth]{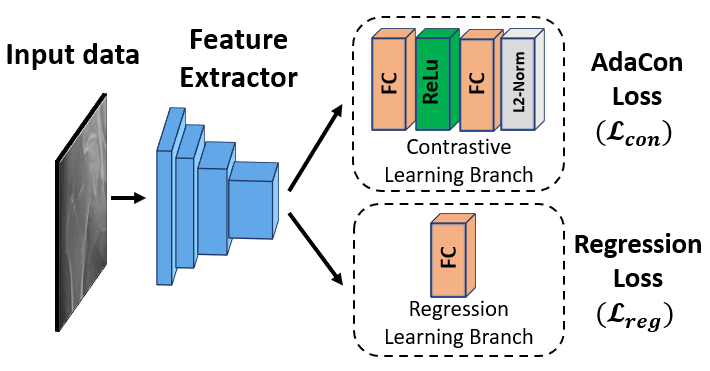}%
\captionsetup{labelfont={color=CLRBlue},font={color=CLRBlue}}
\caption{Under the AdaCon framework, a contrastive prediction branch and a regression branch are added after the feature layer for feature representation learning and regression prediction respectively. The constrastive learning branch is supervised with AdaCon loss which is trained together with a regression loss in a multi-task scheme. 
}
\label{framework_fig}
\end{figure}

The regression branch is used to generate the prediction output for the target value from the input batch $\{x_i; i\in \gI\}$, and can be supervised with either MSE or robust alternatives like L1 and Huber loss. The regression loss, $\gL_{reg}$, is optimized together with the contrastive loss in a multi-task training scheme. The total loss is:
\begin{equation}
     \mathcal{L}_{total} = \gamma_{reg}  \mathcal{L}_{reg} + \gamma_{con}  \mathcal{L}_{con} \wdairv{\:,}
\end{equation}
where $\gamma_{reg}$ and $\gamma_{con}$ are \wdairv{weights} for regression and adaptive contrastive loss \wdairv{respectively and control the relative importance of the two tasks. We fix $\gamma_{reg} = 1$ and set $\gamma_{con}$ such that the weighted loss values for both tasks are within similar magnitudes after the first training epoch. We show in Tables \ref{bmd_gamma} and \ref{lvef_gamma} that our results are robust to slight differences in selected values for the weight parameter. A visual illustration of our framework is shown in Fig. \ref{framework_fig}}

\section{Experiments}
We demonstrate the effectiveness of our method on two medical image regression tasks: BMD estimation from X-ray images and LVEF prediction from echocardiogram videos.

\subsection{BMD Estimation from X-ray Images}
\subsubsection{Experimental setting} \label{bmd_expset}\hfill

\textbf{Dataset.} Dual-energy X-ray absorptiometry (DXA) is the gold standard for diagnosing osteoporosis and monitoring changes in BMD over time.
We use the BMD readings from DXA scans as ground truth labels and collect corresponding X-ray images taken within 6 months of the DXA scans as our model input, following similar settings in ~\cite{zheng2021semi, hsieh2021automated}.
In total, we collect 475 valid hip X-ray images from 317 patients who have had a DXA scan performed between 201\wdairv{6}-2020. \wdairv{The X-ray images were collected using Carestream Health - DRX-EVOLUTION and Canon Inc. - CXDI Control Software NE machines. DXA scans were performed using Hologic - Horizon A machines. We present some  summary statistics of the dataset in Tables \ref{bmd_patstat} and \ref{bmd_machinestat} of Appendix \ref{sec:bmd_sumstat}. The anonymized data was provided by Queen Mary Hospital in Hong Kong. The study was approved by the Institutional Review Board of the University of Hong Kong/Hospital Authority Hong Kong West Cluster (IRB no: UW19-732) and requirement for informed consent was waived.}

\textbf{Preprocessing.} 
\wdairv{The 415 X-ray images were pre-filtered by a radiologist to contain only anteroposterior views of the hip such that the femur-head and femur-neck are clearly visible. Those containing displaced fractures in the femur-neck, where the head has been entirely detached, or containing a prosthesis were also excluded. We then take square crops of the X-ray images around the femur-head and neck region, which are the region of interests corresponding to hip DXA scans. We first use the tool LabelImg\footnote{https://github.com/tzutalin/labelImg } to find the square region that tightly encloses the femur head and neck. We then expand the bounds by 10\% in all directions to incorporate neighbourhood regions within our crop. This yielded a total of 667 square crops centered on the femur with sizes between 800-1200 pixels. Some examples of valid and invalid crops are shown in Fig.~\ref{crop_examppic} of Appendix~\ref{sec:bmd_sumstat}.} 

We apply random data augmentations including horizontal flips, rotations, and frame jitters during sampling. The crops \wdairv{are} resized to $128 \times 128$ pixels before being fed into our network.  
We perform four-fold cross-validation for all the experiments and report the averaged performance \wdairv{across folds.}

\textbf{Training Details.} 
We use EfficientNet-B1~\cite{tan2019efficientnet} as our backbone for BMD estimation and initialize with weights pretrained on ImageNet. 
We use SGD as our optimization algorithm with learning rate \wdairv{of} $10^{-2}$, momentum \wdairv{of} 0.9, and weight decay \wdairv{of} $10^{-4}$.  
The network is trained for 6,000 iterations with learning rate decay of factor 0.1 at 3,000 and 4,500 iterations. 
For the regression prediction branch, we use L1 loss for supervision, which provides the highest regression baseline from our experiments (see Table \ref{loss_comp_reg}). 

Contrastive learning benefits from a large batch size since more negative sample pairs can be formed to improve results.  
To obtain more samples per batch, we enlarge the batch size by following the batch augmentation methodology in~\cite{hoffer2019augment}.
We use an original batch size of 8 and an augmentation multiple of 8. Experiments were conducted with a V100 GPU.

\subsubsection{Comparison with state-of-the-art methods}\label{bmd_expsota} 

Hsieh \textit{et al.} ~\cite{hsieh2021automated} used only L1 loss to directly regress BMD values from femur-neck crops of X-ray images. Zheng \textit{et al.}~\cite{zheng2021semi}, in addition to MSE loss, applied adaptive triplet loss directly on the feature layer without using L2 normalization. Semi-supervised learning with unlabeled data was also used to boost performance. We compare AdaCon with these two methodologies to demonstrate the effectiveness of our framework. To fairly compare with their results, we use the same EfficientNet-B1 backbone.

\begin{table}[h!]
  \caption{Comparison with the state-of-the-art methods for BMD estimation.}
  \label{loss_comp_sota}
  \centering
  \begin{tabular}{c|ccc}
    \hline 
    \textbf{Method}       & \textbf{MAE}$\downarrow$ & \textbf{RMSE}$\downarrow$    & $\mathbf{R}^2 \uparrow$ \\
    \hline
    Hsieh \textit{et al.}~\cite{hsieh2021automated}      & 0.0625    & 0.0789   & 61.38\% \\
    Zheng \textit{et al.}~\cite{zheng2021semi}     & 0.0612    & 0.0773  & 62.04\%       \\
    \textbf{Ours}    & \textbf{0.0592}    & \textbf{0.0759} & \textbf{64.49}\%   \\

    \hline
    
  \end{tabular}

\end{table}

We see from Table \ref{loss_comp_sota} that Zheng \textit{et al.}'s method, leads to better results than Hsieh \textit{et al.}, where only L1 loss is used for training. AdaCon gives us the best performance across all metrics, achieving 0.0592 on MAE, representing a relative improvement of 3.3\% over Zheng \textit{et al.}. \wdairv{In clinical applications, raw BMD values are used to track patient response to osteoporosis treatment or for disease diagnosis~\cite{kranioti2019bone}. Our improved methodology increases accuracy of BMD estimates, which directly translates into more reliable disease tracking and classification applications for computer-aided assessment.  }

\subsubsection{Comparison with other representation learning losses} 
To demonstrate the effectiveness of the proposed adaptive-margin contrastive loss in AdaCon, we compare its performance with alternative loss functions: N-pair loss~\cite{sohn2016improved}, SupCon loss~\cite{khosla2020supervised}, and adaptive triplet loss~\cite{khosla2020supervised}. We also compare with a regression baseline where only a single L1 loss is used. 
For fair comparison, we simply replace our AdaCon loss with these alternative loss functions in our framework and keep other parameters the same. 
For N-Pair and SupCon loss, which are typically used for classification, we treat each unique label as an individual class.

\begin{table}[h!]
  \caption{Comparison of different representation learning loss functions for BMD estimation.}

  \label{loss_comp}
  \centering
  \begin{tabular}{c|ccc}
    \hline 

    \textbf{Method}      & \textbf{MAE}$\downarrow$ & \textbf{RMSE}$\downarrow$  & $\mathbf{R}^2 \uparrow$   \\
    \hline
    N-Pair    & 0.0637    & 0.0816    & 58.70\% \\
    SupCon   & 0.0629    & 0.0794   & 61.15\%   \\
    Regression (baseline)  & 0.0625    & 0.0789   & 61.38\%   \\
    Adaptive Triplet       & 0.0601   & 0.0766    & 63.70\%  \\
    \textbf{Ours}    & \textbf{0.0592}    & \textbf{0.0759}   & \textbf{64.49}\%   \\

    \hline

  \end{tabular}

\end{table}

As shown in Table~\ref{loss_comp}, our method (AdaCon) achieves the best results whereas classification based loss function\wdairv{s} such as N-Pair and SupCon perform worse than the baseline. This is consistent with our expectations since treating all negative samples the same without accounting for label distance can result in inadequate features being learned. 
Adaptive triplet loss lead\wdairv{s} to a decrease in MAE of 0.0024 compared to the regression baseline, but the improvement is still lower than ours. 
AdaCon achieves 0.0592 on MAE, an improvement of 0.0033 over the baseline.

\begin{figure*}[hbt]%
\centering
\begin{subfigure}{0.5 \columnwidth}
\centering
\includegraphics[width=\columnwidth]{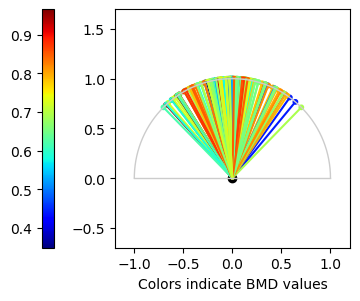}%
\caption{N-pair}%
\label{ang_subfig1}%
\end{subfigure}\hfill%
\begin{subfigure}{0.5 \columnwidth}
\centering
\includegraphics[width=\columnwidth]{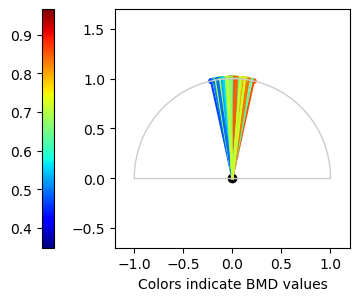}%
\caption{SupCon}%
\label{ang_subfig2}%
\end{subfigure}\hfill%
\begin{subfigure}{0.5 \columnwidth}
\centering
\includegraphics[width=\columnwidth]{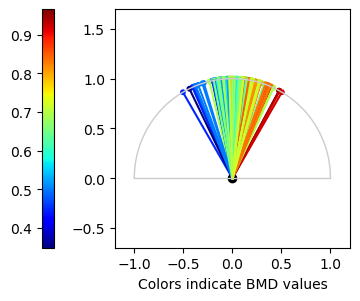}%
\caption{Adaptive Triplet Loss}%
\label{ang_subfig3}%
\end{subfigure}\hfill%
\begin{subfigure}{0.5 \columnwidth}
\centering
\includegraphics[width=\columnwidth]{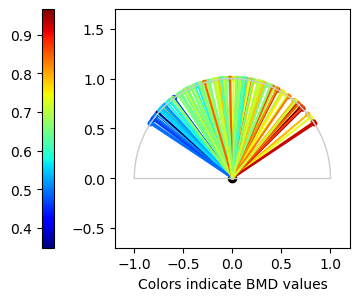}%
\caption{AdaCon}%
\label{ang_subfig4}%
\end{subfigure}\hfill%
\begin{subfigure}{0.5 \columnwidth}
\centering
\includegraphics[width=\columnwidth]{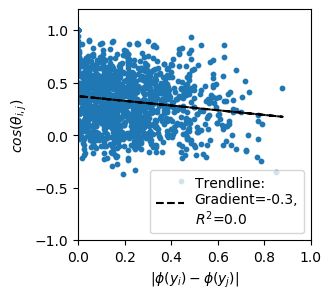}%
\caption{\wdairv{N-pair}}%
\label{scat_subfig1}%
\end{subfigure}\hfill%
\begin{subfigure}{0.5 \columnwidth}
\centering
\includegraphics[width=\columnwidth]{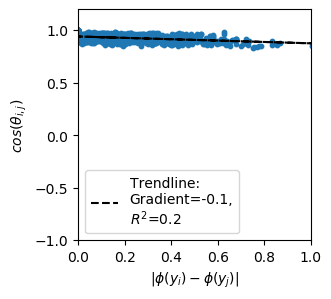}%
\caption{\wdairv{SupCon}}%
\label{scat_subfig2}%
\end{subfigure}\hfill%
\begin{subfigure}{0.5 \columnwidth}
\centering
\includegraphics[width=\columnwidth]{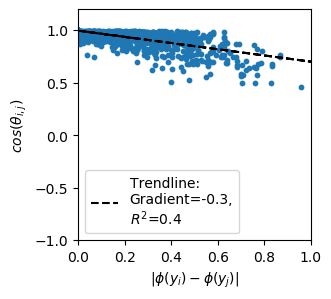}%
\caption{\wdairv{Adaptive Triplet Loss}}%
\label{scat_subfig3}%
\end{subfigure}\hfill%
\begin{subfigure}{0.5 \columnwidth}
\centering
\includegraphics[width=\columnwidth]{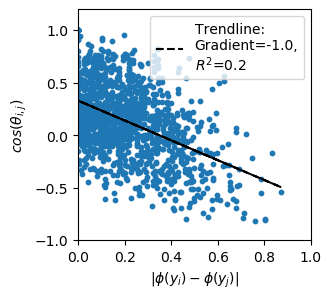}%
\caption{\wdairv{AdaCon}}%
\label{scat_subfig4}%
\end{subfigure}\hfill%

\caption{Plots (a) - (d) are the angular plots of feature projections on the 2d plane. The feature representations of the validation samples are inferred, and the two samples that give the minimum cosine similarity value are used endpoints. The remaining features are distributed based on cosine similarity with the endpoints. 
Blue (red) color represents low (high) BMD. The wider and more orderly the spread of features from low to high BMD, the better the results. 
Plots (e) - (h) are scatter plots for randomly sampled pairs of feature vectors, $i$ and $j$. The x axis is the label distance after applying ECDF; the y axis is feature similarity. The more distinct the downward trend in the scatter plot, the better the results. Fig. \ref{bmd_vis}d and Fig. \ref{bmd_vis}h clearly show AdaCon is able to learn the best feature representations for the BMD task. 
}
\label{bmd_vis}
\end{figure*}

We visualize the learned feature projections in Fig.~\ref{bmd_vis} to better understand the performance improvement. The top row (Fig. \ref{bmd_vis}a-d) shows the angular relationships of the learned feature projections from the validation set, with blue (red) lines representing low (high) BMD values. The bottom row (Fig. \ref{bmd_vis}e-h) plots feature similarity against the absolute difference of $\phi(y)$ for randomly sampled pairs. We can see in Fig. \ref{bmd_vis}d that by using AdaCon, the learned features are well spread out in the angular similarity space, following an orderly progression from low BMD (blue) to high BMD (red). Similarly, we see there is a clear downward trend in Fig. \ref{bmd_vis}h, which demonstrates that pairwise samples with large differences in BMD values tend to have lower cosine similarity. This demonstrates that by using AdaCon, our model is able to learn separable features reflecting label distances based on feature similarities. This helps explain the stronger performance on the regression task. 

On the other hand, we can see from Fig. \ref{bmd_vis}b and Fig. \ref{bmd_vis}c. that feature projections are not well separated using SupCon and adaptive triplet loss. There is also no distinct downward trend in their scatter plots (Fig. \ref{bmd_vis}f and Fig. \ref{bmd_vis}g). Although the features learned by the N-pair loss (Fig. \ref{bmd_vis}a) are well separated, they do not follow the order of low BMD to high BMD. There is also no clear relationship between pair-wise feature similarity and label distance (Fig. \ref{bmd_vis}e). The feature representations learned using these loss functions are therefore not as suitable, leading to worse regression performance.

\subsubsection{Ablation with different regression losses}

AdaCon as a framework can be used as a plug-and-play component with existing regression losses to improve performance. We perform ablation analysis on the BMD dataset with and without AdaCon using three different regression losses: L1, MSE, and Huber loss~\cite{huber1992robust}.
We set the delta parameter for Huber loss at 0.05, which is close to the MAE of our regression baselines.

\begin{table}[h!]
  \caption{Comparison with different regression losses for BMD estimation.}
  \label{loss_comp_reg}
  \centering
  \begin{tabular}{c|ccc}
    \hline
    \textbf{Method}    & \textbf{MAE} $\downarrow$ & \textbf{RMSE} $\downarrow$ &   $\mathbf{R}^2$ $\uparrow$   \\
    \hline
    Huber ($\delta$ = 0.05)      & 0.0823    & 0.1029   &  34.27\% \\
    Huber ($\delta$ = 0.05) + Ours     & 0.0732    & 0.0927     & 46.09\%  \\
    \hline
    MSE    & 0.0719    & 0.0894   &  50.29\%   \\
    MSE + Ours       & 0.0668    & 0.0842    &  56.08\% \\
    \hline
    L1   & 0.0625    & 0.0789        &  61.38\%   \\
    \textbf{L1 + Ours}    & \textbf{0.0592}    & \textbf{0.0759}    &  \textbf{64.49}\%   \\
    \hline
   
  \end{tabular}
\end{table}

We can see from Table \ref{loss_comp_reg} that adding AdaCon consistently boosts performance due to the additional constraints on feature representations introduced through contrastive learning. Relative improvements in MAE of 5.3\%-11.1\% can be seen with the loss convergence greatly benefiting from the multi-task prediction.

\wdairv{

\subsubsection{Ablation with different weight parameters}

During training, we fix $\gamma_{reg} = 1$ and set $\gamma_{con}$ such that the weighted loss values for both tasks are within similar magnitudes after one epoch. We perform ablation analysis on the weight parameter $\gamma_{con}$ using values of 0.0025, 0.0050, and 0,0075 to show that performance is robust to slight differences in chosen value. Results are shown in Table \ref {bmd_gamma}.

\begin{table}[h!]
\captionsetup{labelfont={color=CLRBlue},font={color=CLRBlue}}
  \caption{Comparison with different values of $\gamma_{con}$ for BMD estimation.}
  \label{bmd_gamma}
  \centering
  {\color{CLRBlue}
  \begin{tabular}{c|ccc}
    \hline 
    \textbf{Method}       & \textbf{MAE}$\downarrow$ & \textbf{RMSE}$\downarrow$    & $\mathbf{R}^2 \uparrow$ \\
    \hline
    $\gamma_{reg} = 1, \gamma_{con} = 0.0025$      & 0.0603	& 0.0765 &	63.93\% \\
    $\boldsymbol{\gamma_{reg} = 1, \gamma_{con} = 0.0050}$      & \textbf{0.0592}	& \textbf{0.0759} &	\textbf{64.49}\% \\
    $\gamma_{reg} = 1, \gamma_{con} = 0.0075$      & 0.0596	& 0.0763 &	64.16\% \\

    \hline
  \end{tabular}
  }
\end{table}

Using $\gamma_{con} = 0.0050$ and $\gamma_{con} = 0.0075$ for training gives us similar results. Prediction performance using  $\gamma_{con} = 0.0025$, where less importance is given to contrastive learning, is slightly worse but still outperforms Zheng \textit{et al.}~\cite{zheng2021semi} (0.0603~\vs~0.0612). Overall,  AdaCon achieves superior results for similar magnitudes of $\gamma_{con}$.

\subsubsection{Ablation under different training schemes}

We examine the use of different training schemes on model performance to justify our choice of multi-task training. We compare with a two-stage training scheme consisting of pretraining with contrastive learning and fine-tuning with regression. Pretraining with only AdaCon loss was done using SGD with a learning rate of $10^{-2}$ for a total of 3,000 iterations and learning rate decay of factor 0.1 at 1,500 and 2,250 iterations. Fine-tuning using L1 loss was done with a learning rate of $10^{-3}$ for a total of 3,000 iterations and learning rate decay of 0.1 at 1,500 and 2,250 iterations. Results are shown in Table \ref{train_scheme}.

\begin{table}[h!]
\captionsetup{labelfont={color=CLRBlue},font={color=CLRBlue}}
  \caption{Results using different training schemes for BMD prediction.}
  \label{train_scheme}
  \centering
{\color{CLRBlue}
  \begin{tabular}{c|ccc}
    \hline
    \textbf{Method}      & \textbf{MAE}$\downarrow$ & \textbf{RMSE}$\downarrow$ & $\mathbf{R}^2$ $\uparrow$  \\
    \hline
    N-Pair (two-stage)  &  0.0864 & 0.1140  & 19.76\%\\
    N-Pair (multi-task)      & 0.0637    & 0.0816    & 58.70\%\\
    \hline
    SupCon (two-stage) & 0.0927&0.1207  & 9.13\% \\
    SupCon (multi-task)     & 0.0629    & 0.0794    & 61.15\%  \\
    \hline
    Adaptive Triplet (two-stage) & 0.0691 & 0.0874  & 52.58\%\\
    Adaptive Triplet (multi-task)    & 0.0601    & 0.0766     & 63.70\%   \\
    \hline
    Ours (two-stage) & 0.0964 & 0.1265 & 1.38\% \\
     \textbf{Ours   (multi-task)}    & \textbf{0.0592}    & \textbf{0.0759}  & \textbf{64.49}\%   \\
    \hline
    
  \end{tabular}}
\end{table}

We can see that two-stage training schemes leads to poor convergence due to inadequate network initialization for regression fine-tuning. Multi-task training on the other hand reliably leads to better performance for all cases, with best results seen from using AdaCon with multi-task training.

\subsubsection{External validation}

We validate performance using two additional datasets: one with X-ray images obtained using Fujifilm machines from Queen Mary Hospital (Fujifilm dataset), and another with X-ray images obtained from Chang Gung Memorial Hospital in Taiwan (CGMH dataset). The X-ray images were paired with ground truth BMD readings taken using Hologic - Horizon A machines. 

\textbf{Fujifilm dataset.} 
We perform validation using X-ray images taken by Fujifilm machines, an out of sample machine model, to demonstrate generalization across different machine types. The dataset consists of 19 X-rays from 19 patients collected from Queen Mary Hospital. 
We use the same pre-processing protocol as detailed in Section \ref{bmd_expset}. Summary statistics are shown in Table \ref{bmd_patstat_fuji} of Appendix \ref{sec:bmd_sumstat}. We use the trained models from our four-fold cross validation experiments in Section \ref{bmd_expsota} to run inference, taking the average output of the four models as our final prediction. Results are shown in Table~\ref{bmd_ext_fuji}.

\begin{table}[h!]
\captionsetup{labelfont={color=CLRBlue},font={color=CLRBlue}}
  \caption{Test results for BMD estimation using Fujifilm X-ray machines.}
  \label{bmd_ext_fuji}
  \centering
  {\color{CLRBlue}
  \begin{tabular}{c|ccc}
    \hline 
    \textbf{Method}       & \textbf{MAE}$\downarrow$ & \textbf{RMSE}$\downarrow$    & $\mathbf{R}^2 \uparrow$ \\
    \hline
    Hsieh \textit{et al.}~\cite{hsieh2021automated}      & 0.0684	& 0.0888 &	53.43\% \\
    Zheng \textit{et al.}~\cite{zheng2021semi}     & 0.0651 &	0.0870 &	55.39\%       \\
    \textbf{Ours}    & \textbf{0.0580}    & \textbf{0.0814} & \textbf{60.98}\%   \\

    \hline
    
  \end{tabular}
  }

\end{table}

AdaCon (ours) gives the best performance on this validation dataset, which is consistent with our results in Table~\ref{loss_comp_sota}. MAE for AdaCon is 0.0071 lower than the methodology in Zheng \textit{et al}. \cite{zheng2021semi}, representing a relative decrease of 10.9\%.
We also see that the performance metrics are within similar range of the values in Table~\ref{loss_comp_sota} (0.0580 \vs 0.0592).
The results demonstrate that our trained model generalizes well across different X-ray machine manufacturers and that AdaCon learns more relevant features compared to alternative methods.

\textbf{CGMH dataset.} We perform a second external validation using data provided by Chang Gung Memorial Hospital  (CGMH) to demonstrate generalization across different hospitals. The dataset consists of 100 X-ray images paired with BMD labels. A number of X-rays were excluded based on our pre-processing protocol as they contained displaced fractures or were not of anteroposterior views as determined by a radiologist. The remaining dataset consisted of 61 X-rays from 61 individual patients. Patient summary statistics are shown in Table \ref{bmd_patstat_cgmh} of Appendix \ref{sec:bmd_sumstat}. Results are shown in Table \ref{bmd_ext_cgmh}.

\begin{table}[h!]
\captionsetup{labelfont={color=CLRBlue},font={color=CLRBlue}}
  \caption{Test results for BMD estimation using CGMH dataset.}
  \label{bmd_ext_cgmh}
  \centering
  {\color{CLRBlue}
  \begin{tabular}{c|ccc}
    \hline 
    \textbf{Method}       & \textbf{MAE}$\downarrow$ & \textbf{RMSE}$\downarrow$    & $\mathbf{R}^2 \uparrow$ \\
    \hline
    Hsieh \textit{et al.}~\cite{hsieh2021automated}      & 0.0748	& 0.0988 &	59.10\% \\
    Zheng \textit{et al.}~\cite{zheng2021semi}     & 0.0741	& 0.0980 &	59.77\%       \\
    \textbf{Ours}    & \textbf{0.0722}    & \textbf{0.0962} & \textbf{61.19}\%   \\

    \hline
  \end{tabular}
  }
\end{table}

The MAE values based on our methodology is higher than what we obtained in Table 1 (0.0722 \vs 0.0592), but the standard deviation of BMD in the CGMH dataset is also larger, as seen in Tables \ref{bmd_patstat_cgmh} and \ref{bmd_patstat} (0.1568 \vs 0.1291). 
Our proposed methodology performs better than both Hsieh \textit{et al.} and Zheng \textit{et al.}, and decreases MAE by 0.0019 compared to Zheng \textit{et al.}, a relative improvement of 3\%. These results show that our method generalizes across datasets from different hospitals. 

}

\subsection{LVEF Prediction}

\subsubsection{Experimental Setting} \hfill

\textbf{Dataset.} We use the EchoNet-Dynamic dataset, which is a public dataset consisting of 10,036 apical-4-chamber echocardiogram videos collected from Stanford University Hospital~\cite{ouyang2020video}. The dataset is paired with left-ventricular ejection fraction (LVEF), end-systolic volume (ESV) and end-diastolic volume (EDV) labels. \wdairv{The echocardiogram videos are collected using iE33, Sonos, Acuson SC2000, Epiq 5G, and Epiq 7C ultrasound machines. Summary statistics on the dataset are shown in Table \ref{lvef_stat} of Appendix \ref{sec:bmd_sumstat}. Additional details of the dataset can also be found in \cite{ouyang2019echonet}.}
The data has already been pre-processed by removing irrelevant text information and re-scaled to $112 \times 112$ pixels. Training, validation, and test splits have already been prepared. Frame jitter is applied as part of the augmentation process.

\textbf{Training Details.} We follow the original model implementation and training scheme specified in~\cite{ouyang2020video} and add our contrastive prediction branch after the feature layer. The backbone uses a R2+1D ResNet~\cite{tran2018closer} \wdairv{pretrained on Kinetics-400~\cite{kay2017kinetics}} and the entire model is trained with SGD \wdairv{using a} learning rate of $10^{-4}$ and momentum of 0.9 over 45 epochs\wdairv{,} with decay of factor 0.1 every 15 epochs. MSE is used as the regression loss function. Clips of 32 frames, sampled from the video at a rate of 1 in every 2 frames, were used as \wdairv{the} model input \wdairv{with a} batch size of 20. Due to the large memory requirements, we do not add additional batch augmentation operations. We use the validation set to choose the best trained model over all epochs and report results on the test splits. Each experiment was run on \wdairv{three} V100 \wdairv{GPUs} for approximately 12 hours.

\subsubsection{Comparison with state-of-the-art methods} 

State-of-the-art performance on the Echonet-Dynamic dataset is achieved by Ouyang \textit{et al.}~\cite{ouyang2020video}, where sampled frames are regressed directly on LVEF values using a R2+1D backbone and MSE loss. We compare our performance using AdaCon with their results in Table \ref{loss_comp_sota_lvef}.

\begin{table}[h!]
  \caption{Comparison with the state-of-the-art methods for LVEF prediction.}

  \label{loss_comp_sota_lvef}
  \centering
  \begin{tabular}{c|ccc}
    \hline 
    \textbf{Method}       & \textbf{MAE}$\downarrow$ & \textbf{RMSE}$\downarrow$    & $\mathbf{R}^2 \uparrow$ \\
    \hline
    Ouyang \textit{et al.}~\cite{ouyang2020video}      & 4.10    & 5.40   & 80.5\% \\
    \textbf{Ours}    & \textbf{3.86}    & \textbf{5.07} & \textbf{82.8}\%   \\

    \hline
    \multicolumn{4}{c} 
    
  \end{tabular}

\end{table}

We can see that AdaCon out-performs Ouyang \textit{et al.}'s methodology and decreases MAE by 0.24, a relative improvement of 5.9\%. \wdairv{LVEF is commonly used as an indicator to diagnose heart conditions~\cite{maeder2009heart}, and more accurate predictions using our methodology can lead to more reliable tracking and diagnosis of cardiac disease. }

\subsubsection{Comparison with other representation learning losses}

\begin{figure*}[h!]%
\centering
\begin{subfigure}{0.5 \columnwidth}
\centering
\includegraphics[width=\columnwidth]{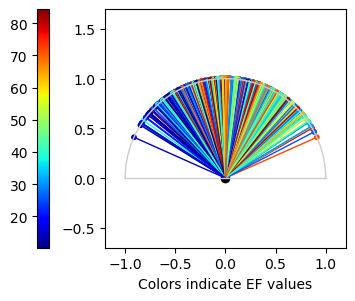}%
\caption{N-pair}%
\label{angb_fig1}%
\end{subfigure}\hfill%
\begin{subfigure}{0.5 \columnwidth}
\centering
\includegraphics[width=\columnwidth]{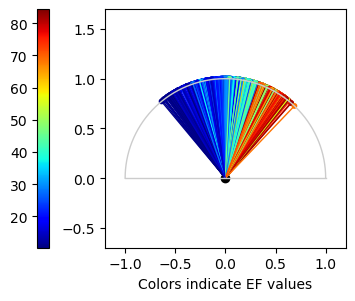}%
\caption{SupCon}%
\label{angb_fig2}%
\end{subfigure}\hfill%
\begin{subfigure}{0.5 \columnwidth}
\centering
\includegraphics[width=\columnwidth]{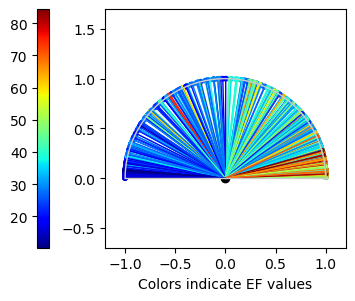}%
\caption{Adaptive Triplet Loss}%
\label{angb_fig3}%
\end{subfigure}\hfill%
\begin{subfigure}{0.5 \columnwidth}
\centering
\includegraphics[width=\columnwidth]{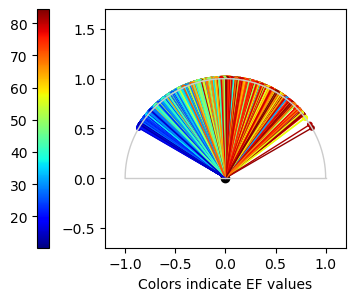}%
\caption{AdaCon}%
\label{angb_fig4}%
\end{subfigure}\hfill%
\begin{subfigure}{0.5 \columnwidth}
\centering
\includegraphics[width=\columnwidth]{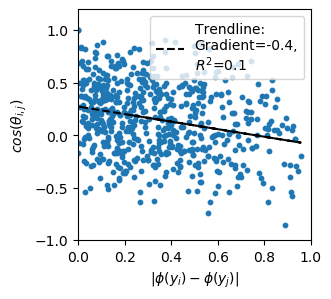}%
\caption{\wdairv{N-pair}}%
\label{scatb_fig1}%
\end{subfigure}\hfill%
\begin{subfigure}{0.5 \columnwidth}
\centering
\includegraphics[width=\columnwidth]{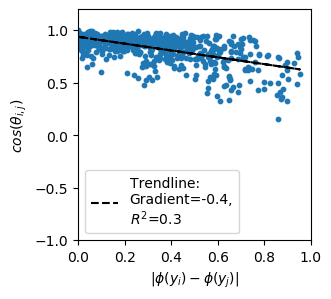}%
\caption{\wdairv{SupCon}}%
\label{scatb_fig2}%
\end{subfigure}\hfill%
\begin{subfigure}{0.5 \columnwidth}
\centering
\includegraphics[width=\columnwidth]{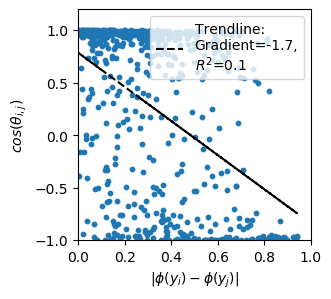}%
\caption{\wdairv{Adaptive Triplet Loss}}%
\label{scatb_fig3}%
\end{subfigure}\hfill%
\begin{subfigure}{0.5 \columnwidth}
\centering
\includegraphics[width=\columnwidth]{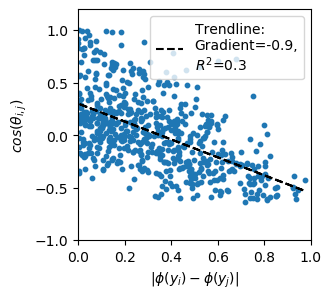}%
\caption{\wdairv{AdaCon}}%
\label{scatb_fig4}%
\end{subfigure}\hfill%

\caption{Plots (a) - (d) are the angular plots of feature projections on the 2d plane. Blue (red) color represents low (high) LVEF values. Plots (e) - (h) are scatter plots for randomly sampled pairs of feature vectors, $i$ and $j$. The x axis is the label distance after applying ECDF; the y axis is feature similarity. Fig. \ref{ef_vis}d and Fig. \ref{ef_vis}h clearly show AdaCon is able to learn the best feature representations for the LVEF task. }
\label{ef_vis}
\end{figure*}

Similar to the BMD estimation task, we compare our proposed loss function against alternatives to demonstrate its effectiveness. We compare against N-pair loss~\cite{sohn2016improved}, SupCon loss~\cite{khosla2020supervised} adaptive triplet loss~\cite{zheng2021semi}, and a baseline where only regression loss is used. Results are reported in Table \ref{loss_comp2}. 

\begin{table}[htb!]

  \caption{Comparison of different representation learning loss functions for LVEF prediction. }
  \label{loss_comp2}
  \centering
  \begin{tabular}{c|ccc}
    \hline
    \textbf{Method}       & \textbf{MAE} $\downarrow$ & \textbf{RMSE} $\downarrow$ & $\mathbf{R}^2$ $\uparrow$ \\
    \hline
    N-Pair      & 4.19    & 5.66 & 78.6\% \\
    Regression (baseline)     & 4.10    & 5.40  & 80.5\%  \\
    SupCon    & 4.05    & 5.37   & 80.7\%   \\
    Adaptive Triplet     & 3.92    & 5.21 & 81.8\%  \\
    \textbf{Ours}        & \textbf{3.86}   & \textbf{5.07} & \textbf{82.8}\% \\
    \hline
    \multicolumn{4}{c} 
    
  \end{tabular}
\end{table}

We see that using N-Pair loss leads to reduced performance when compared to baseline, whilst SupCon and adaptive triplet loss decreased MAE by 0.05 and 0.18 respectively. The largest performance gain occurs when using AdaCon, achieving MAE of 3.86, or an improvement of 0.24 over baseline.

We visualize the learned feature representations in Fig. \ref{ef_vis} and see that by using AdaCon for feature representation learning, the feature vectors are able to achieve an even and orderly spread in terms of their angular similarity (Fig. \ref{ef_vis}d). A clear downward trend can also be seen between pair-wise feature cosine similarity and label distance, indicating the learned features successfully reflects label order and distance in terms of their similarity values. 

On the other hand, we can see from Fig. \ref{ef_vis}a and Fig. \ref{ef_vis}e that there is little relation between feature similarity and label distance in the representations learned using N-pair loss.
For SupCon, we see a downward-sloping trendline in Fig. \ref{ef_vis}f, but the relationship is not very strong. 
The features learned using adaptive triplet loss are well spread out in Fig. \ref{ef_vis}c, but similarity values are clustered around 1 and -1 with little observations in-between (Fig. \ref{ef_vis}g). Features learned by these alternative loss functions therefore do not reflect label relationships as well as AdaCon.

\subsubsection{Ablation under different regression losses}

We demonstrate that applying AdaCon consistently boosts performance independent of different choices for the regression prediction \wdairv{branch}. We use three different regression loss functions: MSE, L1, and Huber Loss and present the results with and without using AdaCon in Table \ref{loss_comp2_EF}. We set the delta value for Huber loss at 4, which is close to the MAE of our regression baselines. 

\begin{table}[htb!]
  \caption{Comparison with different regression losses for LVEF prediction.} 
  \label{loss_comp2_EF}
  \centering
  \begin{tabular}{c|ccc}
    \hline
    \textbf{Method}         & \textbf{MAE} $\downarrow$ & \textbf{RMSE} $\downarrow$    & $\mathbf{R}^2$ $\uparrow$\\
    \hline
    L1     & 4.51    & 6.22  & 74.1\%  \\
    L1 + Ours       & 4.46    & 6.08    & 74.8\% \\
    
    \hline

    Huber ($\delta = 4$)     & 4.27    & 5.80    & 77.5\%\\
    Huber ($\delta = 4$) + Ours  & 4.24    & 5.74 &   77.9\% \\
    \hline
    MSE       & 4.10    & 5.40   & 80.5\% \\
    \textbf{MSE + Ours}        & \textbf{3.86}   & \textbf{5.07}  & \textbf{82.8}\% \\
    \hline

  \end{tabular} 
\end{table}

Similar to what was observed for BMD estimation, the use of our AdaCon framework consistently boosts performance. Relative improvements in MAE of between 0.7\% - 5.9\% can be seen independent of the regression loss used. The improvement in feature representation learning from the contrastive prediction task reliably improves performance on the regression branch.

\wdairv{

\subsubsection{Ablation with different weight parameters}

We perform ablation analysis on the weight parameter $\gamma_{con}$ using values of 0.50, 0.75, and 1.00 and show results in Table \ref {lvef_gamma}.

\begin{table}[h!]
\captionsetup{labelfont={color=CLRBlue},font={color=CLRBlue}}
  \caption{Comparison with different values of $\gamma_{con}$ for LVEF prediction.}
  \label{lvef_gamma}
  \centering
  {\color{CLRBlue}
  \begin{tabular}{c|ccc}
    \hline 
    \textbf{Method}       & \textbf{MAE}$\downarrow$ & \textbf{RMSE}$\downarrow$    & $\mathbf{R}^2 \uparrow$ \\
    \hline
    $\gamma_{reg} = 1, \gamma_{con} = 0.50$      & 3.95	& 5.20 &	82.0\% \\
    $\boldsymbol{\gamma_{reg} = 1, \gamma_{con} = 0.75}$      & \textbf{3.86}	& \textbf{5.07} &	\textbf{82.8}\% \\
    $\gamma_{reg} = 1, \gamma_{con} = 1.00$      & 3.89	& 5.15 &	82.2\% \\

    \hline
  \end{tabular}
  }
\end{table}

The performance metrics are within the same range for all three weight values and outperform the methodology in Ouyang \textit{et al.} \cite{ouyang2020video}, demonstrating robustness to values of $\gamma_{con}$ within similar magnitudes.

\subsubsection{Ablation under different training schemes}
Pretraining was done by training the model with only AdaCon loss for 25 epochs, using a learning rate of $10^{-4}$, momentum of 0.9, and decay of factor 0.1 at epochs 10 and 20. The model is then fine-tuned with MSE loss by training for 25 additional epochs, starting with a learning rate of $10^{-5}$ and decaying with factor 0.1 at epochs 10 and 20.

\begin{table}[htb!]
\captionsetup{labelfont={color=CLRBlue},font={color=CLRBlue}}
  \caption{Results using different training schemes for LVEF prediction.}
  \label{loss_train_ef_stage}
  \centering
{\color{CLRBlue}
  \begin{tabular}{c|ccc}
    \hline
     \textbf{Method}     & \textbf{MAE} $\downarrow$ & \textbf{RMSE} $\downarrow$  & $\mathbf{R}^2$ $\uparrow$  \\
    \hline
     N-Pair (two-stage)  & 4.48 & 6.06 &  75.4\% \\
     N-Pair (multi-task)      & 4.19    & 5.66  & 78.6\%\\
    \hline
     SupCon (two-stage)   & 4.46 & 6.06  & 75.5\%\\
     SupCon (multi-task)     & 4.05    & 5.37  & 80.7\%  \\
    \hline
     Adaptive Triplet (two-stage)  & 4.33 & 5.86  & 77.1\% \\
     Adaptive Triplet (multi-task)    & 3.92    & 5.21 & 81.8\% \\
    \hline
     Ours (two-stage) & 4.15&5.52  & 79.6\% \\
     \textbf{Ours (multi-task)}    & \textbf{3.86}   & \textbf{5.07}  & \textbf{82.8}\% \\

    \hline
    
  \end{tabular}} \\

\end{table}

We see from Table \ref{loss_train_ef_stage} that multi-task training is consistently more effective than two-stage training. Although the difference between the two schemes is smaller than as seen with the BMD dataset (Table \ref{train_scheme}), the two-stage training still fails to out-perform the plain regression baseline. Overall, the best performance is observed when using AdaCon with muti-task training.

\subsubsection{External validation}
We perform external validation on the CAMUS echocardiogram sequence dataset ~\cite{leclerc2019deep} to demonstrate generalization of LVEF prediction across hospitals. The CAMUS dataset contains a series of echocardiogram sequences taken from 500 patients at the University Hospital of St Etienne in France and is split into a training set of 450 patients and a test set of 50 patients. The dataset is acquired using GE Vivid E95 ultrasound scanners and consists of a 2-chamber sequence view and a 4-chamber sequence view for each patient. Manual segmentations and volume labels for the EDV and ESV frame are provided for the training set only. Sequences are also classified into three classes of quality: good, medium, or poor. The dataset was originally designed for a segmentation challenge using the EDV and ESV frames, but the sequence data and labels can be easily adapted to perform LVEF prediction. 

We evaluate performance of our trained models using the 4-chamber training set sequences. Table \ref{lvef_camus} in Appendix \ref{sec:bmd_sumstat} shows summary statistics of the dataset. In addition to differences in acquisition machine, the sequences in the CAMUS dataset also contain much fewer frames on average (20 frames \vs{} 175 frames). We resize all sequences to 112 × 112 pixels and use the same parameters and data processing pipeline. We show results on overall performance as well as performance categorized by sequence quality in Table \ref{loss_ext_camus}.

\begin{table}[h!]
\captionsetup{labelfont={color=CLRBlue},font={color=CLRBlue}}
  \caption{Test results for LVEF prediction on CAMUS dataset based on quality.}
  \label{loss_ext_camus}
  \centering
  {\color{CLRBlue}
  \begin{tabular}{c|c|ccc}
    \hline 
    \textbf{Quality}  & \textbf{Method}       & \textbf{MAE}$\downarrow$ & \textbf{RMSE}$\downarrow$    & $\mathbf{R}^2 \uparrow$ \\
    \hline
    Good
    & Ouyang \textit{et al.}~\cite{ouyang2020video}      & 6.18	& 7.91 &	49.9\% \\
     & \textbf{Ours}    & \textbf{5.95}    & \textbf{7.41} & \textbf{56.1}\%   \\

    \hline
    Medium & Ouyang \textit{et al.}~\cite{ouyang2020video}      & 7.96	& 10.33	& 38.1\% \\
     & \textbf{Ours}    & \textbf{7.41}    & \textbf{9.60} & \textbf{46.5}\%   \\

    \hline
    Poor & Ouyang \textit{et al.}~\cite{ouyang2020video}      & 11.20 &	13.09 &	-36.6\% \\
     & \textbf{Ours}    & \textbf{9.68}    & \textbf{11.96} & \textbf{-14.2}\%   \\

    \hline
    \hline
    Overall & Ouyang \textit{et al.}~\cite{ouyang2020video}      & 7.23 &	9.34 &	40.10\% \\
     & \textbf{Ours}    & \textbf{6.78}    & \textbf{8.68} & \textbf{48.22}\%   \\

    \hline
    \multicolumn{4}{c} 
    
  \end{tabular}
  }
\end{table}

Overall performance shows that AdaCon (ours) is better than Ouyang \textit{et al.} and decreases MAE by 0.45, a relative improvement of 6\%. MAE is higher than on EchoNet-Dynamic results in Table \ref{loss_comp_sota_lvef} however (6.78 \vs 3.86) because CAMUS sequences have much fewer frames on average. We also see that our methodology performs consistently better in all three quality classes compared to the model trained only using regression loss. The use of our AdaCon framework for improved feature representation learning consistently leads to stronger performance, even across different data sources. 

}

\wdairv{
\section{Discussion}

We demonstrated the effectiveness of our AdaCon loss function through ablation experiments in Tables \ref{loss_comp} and \ref{loss_comp2}. We also discussed several desirable properties of our adaptive margin function in Section \ref{adacon_method}. Although we achieve good empirical results, there are still possible limitations to our current formulation that could potentially be improved. The current adaptive margin function imposes a prior assumption on the mapping between label space distance and feature space distance, which may not be the most optimal. It is possible that a learned mapping or one with looser constraints may perform better. Also, the margin function relies on the ECDF for transformation, which implicitly assumes that the sample label distribution of the training set is representative of true label distribution. This may be reasonable for relatively large datasets but may not be the case for smaller datasets or those that have been artificially skewed. Thus, there may exist alternative formulations of the adaptive margin function that lead to better performance. The task of finding a better adaptive margin function is a potential direction for future research. 
}

\section{Conclusion}

We present a novel adaptive contrastive learning framework, AdaCon, for image regression in computer-aided disease assessment. The adaptive-margin loss function used by AdaCon allows us to capture label ordering and probability-normalized distance relationships in our learned features. The features better reflect continuous relationships between sample labels, thus improving performance when applied to regression tasks. 
The effectiveness of our method has been demonstrated on two medical image regression tasks: BMD estimation from X-ray images and LVEF prediction from echocardiogram videos. \wdairv{Results are also consistent when tested on external validation datasets for both tasks.}
Our method is general and can be flexibly combined with existing regression loss functions, such as MSE and L1, as well as different feature-extraction backbones. \wdairv{This improved methodology can help increase accuracy and reliability of deep regression tasks for various medical applications.  }

\wdairv{
\section{Acknowledgement}

This work was partially supported by a research grant from Shenzhen Municipal Central Government Guides Local Science and Technology Development Special Funded Projects (2021Szvup139) and a research grant from Hong Kong Research Grants Council (RGC) under the National Nature Science Foundation of China/Research Grants Council Joint Research Scheme (N\_HKUST627/20). The authors would like to thank Dr. FANG Xin Hao, Benjamin from Queen Mary Hospital, Alistair Yap, and Richard Du for their help in obtaining the BMD dataset. We also thank the Chang Gung Memorial Hospital and Dr. KUO Chang-Fu for kindly making their data available for external validation use.
}

\bibliographystyle{IEEEtranN}
\small{\bibliography{refs}}

\clearpage

\appendices

\setcounter{table}{0}
\renewcommand{\thetable}{C\arabic{table}}
\setcounter{figure}{0}
\renewcommand{\thefigure}{C\arabic{figure}}
\setcounter{equation}{8}

\wdairv{

\section{Derivation of AdaCon Loss}
\label{sec:ada_derive}
We wish to incorporate adaptive margins as defined in (\ref{eqn:adaconstrain}) within the decision boundaries of the standard SupCon loss function, (\ref{eqn:supcon}), for positive pair classification. 

We first consider the case of a batch with three samples: anchor sample $i$, positive sample $p$, and negative sample $q$. Under SupCon loss, the sigmoid function for identifying the positive pair is:

\begin{equation}
    \frac{{\rm exp}(s \: {\cos(\theta_{i,p})} )}{{\rm exp}(s \: \cos(\theta_{i,p})) + {\rm exp}(s \: \cos(\theta_{i,q})) } \:.
\end{equation}
The positive pair is correctly classified if:

\begin{equation}
    {\rm exp}(s \: \cos(\theta_{i,p})) > {\rm exp}(s \: \cos(\theta_{i,q})) \:,
\end{equation}
or equivalently:

\begin{equation}
     \cos(\theta_{i,p}) > \cos(\theta_{i,q}) \:.
\end{equation}

We incorporate our constraint in (\ref{eqn:adaconstrain}) into our decision boundary, such that the positive pair is correctly classified only if $\cos(\theta_{i,p}) > \cos(\theta_{i,q}) + d_{i,q}$. This gives us the modified sigmoid:

\begin{equation}
    \frac{{\rm exp}(s \: {\cos(\theta_{i,p})} )}{{\rm exp}(s \: \cos(\theta_{i,p})) + {\rm exp}(s (\cos(\theta_{i,q}) + d_{i,q})) } \:,
\end{equation}
and the corresponding cross-entropy loss:

\begin{equation}
    - \log \frac{{\rm exp}(s \: {\cos(\theta_{i,p})} )}{{\rm exp}(s \: \cos(\theta_{i,p})) + {\rm exp}(s (\cos(\theta_{i,q}) + d_{i,q})) } \:,
\end{equation}

Extending this to an arbitrary number of positive and negative pairs in the sample batch gives us:

\begin{equation}
    \sum_{i\in \gI} \frac{-1}{|\gP (i)|} \sum_{p \in \gP (i)} {\rm log} \frac{{\rm exp}(s \: \cos(\theta_{i,p}) )}{\sum_{a \in \gI \setminus i} {\rm exp}(s(\cos(\theta_{i,a}) + d_{i,a}))} 
\end{equation}
where by definition of our adaptive margin in (\ref{eqn:adamargin}), $d_{i,a} = 0$ if $a \in \gP (i)$. This gives us our final AdaCon loss function in (\ref{eqn:adacon_loss}). 
}

\section{AdaCon loss with \wdairv{one} positive and \wdairv{one} negative Pair}
\label{sec:ada_1pair}

We show that the AdaCon loss function can be seen as a general formulation for \wdairv{distance} metric learning with variable margins and reduces to approximate adaptive triplet loss when using \wdairv{one} positive and \wdairv{one} negative pair. Making use of Taylor approximations for $\rm log$ and $\rm exp$, we can see:

\[ \gL_{con} = - {\rm log} \frac{{\rm exp}(s(z_i^T z_p + d_{i,p}))}{{\rm exp}(s(z_i^T z_p + d_{i,p})) + {\rm exp}(s(z_i^T z_n + d_{i,n}))}  \]
\[ ={\rm log} ( 1+ {\rm exp}(s(z_i^T z_n - z_i^T z_p + d_{i,n})) ) \]
\[\approx {\rm exp}(s(z_i^T z_n - z_i^T z_p + d_{i,n})) \]
\[\approx 1 + s(z_i^T z_n - z_i^T z_p) + sd_{i,n} \]
\[= 1  + s d_{i,n} - \frac{s}{2}(\|z_i - z_n\|^2 - \|z_i - z_p\|^2) \]
\[= \frac{s}{2} \{ \|z_i - z_p\|^2 - \|z_i - z_n\|^2  + 2d_{i,n}\} + 1 \]
\[ \propto \|z_i - z_p\|^2 - \|z_i - z_n\|^2  + 2d_{i,n} \wdairv{\:.} \] 
The final line of the derivation approximates the adaptive triplet loss function~\cite{zheng2021semi} where the margin value $d_{i,n}$ varies depending on the labels of the sample pairs.

\wdairv{

\section{Details on datasets}
\label{sec:bmd_sumstat}

\begin{table}[htb!]
\captionsetup{labelfont={color=CLRBlue},font={color=CLRBlue}}
  \caption{Patient statistics for BMD dataset.}
  \label{bmd_patstat}
  \centering
  {\color{CLRBlue}
  \begin{tabular}{c|c}
    \hline
    \textbf{Metric}         & \textbf{Value}\\
    \hline
    Total number of patients     & 317  \\
    Females (\%)     & 280 (88.0\%)   \\
    Mean age (sd), years      & 74.79 (11.95)   \\
    Mean BMD (sd), g/$cm^2$  & 0.6968 (0.1291) \\
    Average no. of films per patient & 1.31 \\
    \hline
  \end{tabular} 
  }
\end{table}

\begin{table}[htb!]
\captionsetup{labelfont={color=CLRBlue},font={color=CLRBlue}}
  \caption{Manufacturer information for X-rays in BMD dataset.}
  \label{bmd_machinestat}
  \centering
  {\color{CLRBlue}
  \begin{tabular}{c|c}
    \hline
    \textbf{Manufacturer - Model }         & \textbf{Number (\% of total)}\\
    \hline
    Canon Inc. - CXDI Control Software NE   & 176 (37\%)  \\
    Carestream Health - DRX-EVOLUTION     & 299 (63\%)   \\
    \hline
    Total & 475 (100\%) \\
    \hline
  \end{tabular} 
  }
\end{table}

\begin{table}[h!]
\captionsetup{labelfont={color=CLRBlue},font={color=CLRBlue}}
  \caption{Patient statistics for Fujifilm dataset.}
  \label{bmd_patstat_fuji}
  \centering
  {\color{CLRBlue}
  \begin{tabular}{c|c}
    \hline
    \textbf{Metric}         & \textbf{Value}\\
    \hline
    Total number of patients     & 19  \\
    Females (\%)     & 14 (73.6\%)   \\
    Mean age (sd), years      & 68.6 (14.52)   \\
    Mean BMD (sd), g/$cm^2$  & 0.7149 (0.1303) \\
    \hline
  \end{tabular} 
  }
\end{table}

{\color{red}\begin{table}[h!]
\captionsetup{labelfont={color=CLRBlue},font={color=CLRBlue}}
  \caption{Patient statistics for CGMH dataset.}
  \label{bmd_patstat_cgmh}
  \centering
  {\color{CLRBlue}
  \begin{tabular}{c|c}
    \hline
    \textbf{Metric}         & \textbf{Value}\\
    \hline
    Total number of patients     & 61  \\
    Females (\%)     & 48 (78.7\%)   \\
    Mean age (sd), years      & 68.95 (12.14)   \\
    Mean BMD (sd), g/$cm^2$  & 0.7325  (0.1568) \\
    \hline
  \end{tabular} 
  }
\end{table}}

\begin{table}[h!]
\captionsetup{labelfont={color=CLRBlue},font={color=CLRBlue}}
  \caption{Summary statistics for LVEF dataset}
  \label{lvef_stat}
  \centering
  {\color{CLRBlue}
  \begin{tabular}{c|c}
    \hline
    \textbf{Metric}         & \textbf{Value}\\
    \hline
    Total number of patients     & 10,036  \\
    Average number of frames (sd) & 175 (57) \\
    Average frames per second (sd) & 50.9 (6.8) \\
    Mean LVEF (sd), \%     & 55.75 (12.37)   \\
    \hline
  \end{tabular} 
  }
\end{table}

\begin{table}[h!]
\captionsetup{labelfont={color=CLRBlue},font={color=CLRBlue}}
  \caption{Summary statistics for CAMUS dataset.}
  \label{lvef_camus}
  \centering
  {\color{CLRBlue}
  \begin{tabular}{c|c}
    \hline
    \textbf{Metric}         & \textbf{Value}\\
    \hline
    Total number of patients     & 450  \\
    Average number of frames (sd) & 20 (4) \\
    Video quality & Good: 260, Medium: 148, Poor: 42 \\
    Mean LVEF (sd), \%     & 52.03 (12.07)   \\
    \hline
  \end{tabular} 
  }
\end{table}

\begin{figure*}[b!]%
\centering
\begin{subfigure}{1.9 \columnwidth}
\centering
\includegraphics[width=\columnwidth]{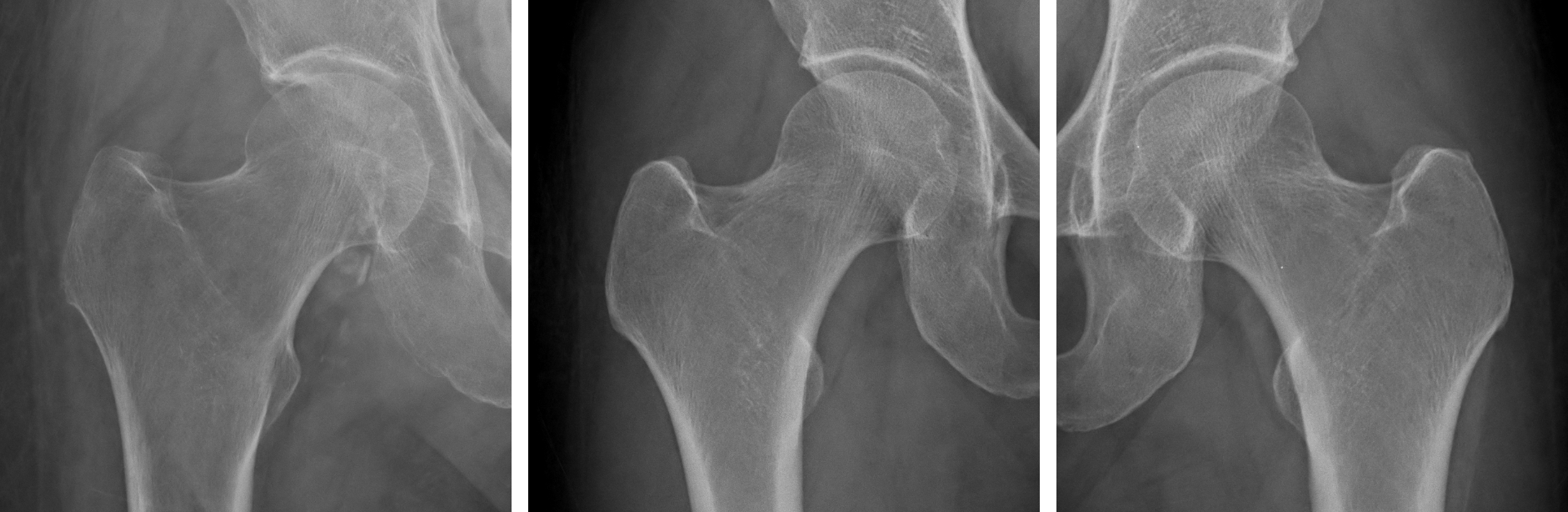}%
\captionsetup{labelfont={color=CLRBlue},font={color=CLRBlue}}
\caption{Examples of valid crops that are used in the training dataset}%
\label{crop_good}%
\end{subfigure}\hfill%

\centering
\begin{subfigure}{1.9 \columnwidth}
\centering
\includegraphics[width=\columnwidth]{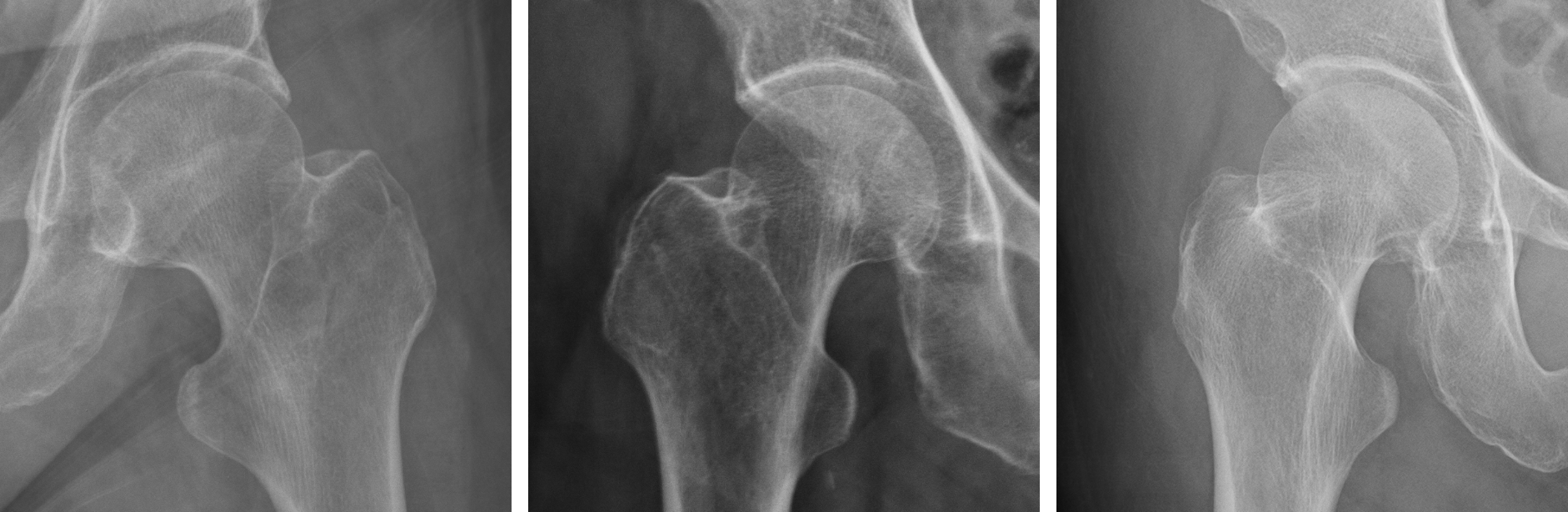}%
\captionsetup{labelfont={color=CLRBlue},font={color=CLRBlue}}
\caption{Examples of crops not of anteroposterior view that have been excluded}%
\label{crop_nonap}%
\end{subfigure}\hfill%

\centering
\begin{subfigure}{1.9 \columnwidth}
\centering
\includegraphics[width=\columnwidth]{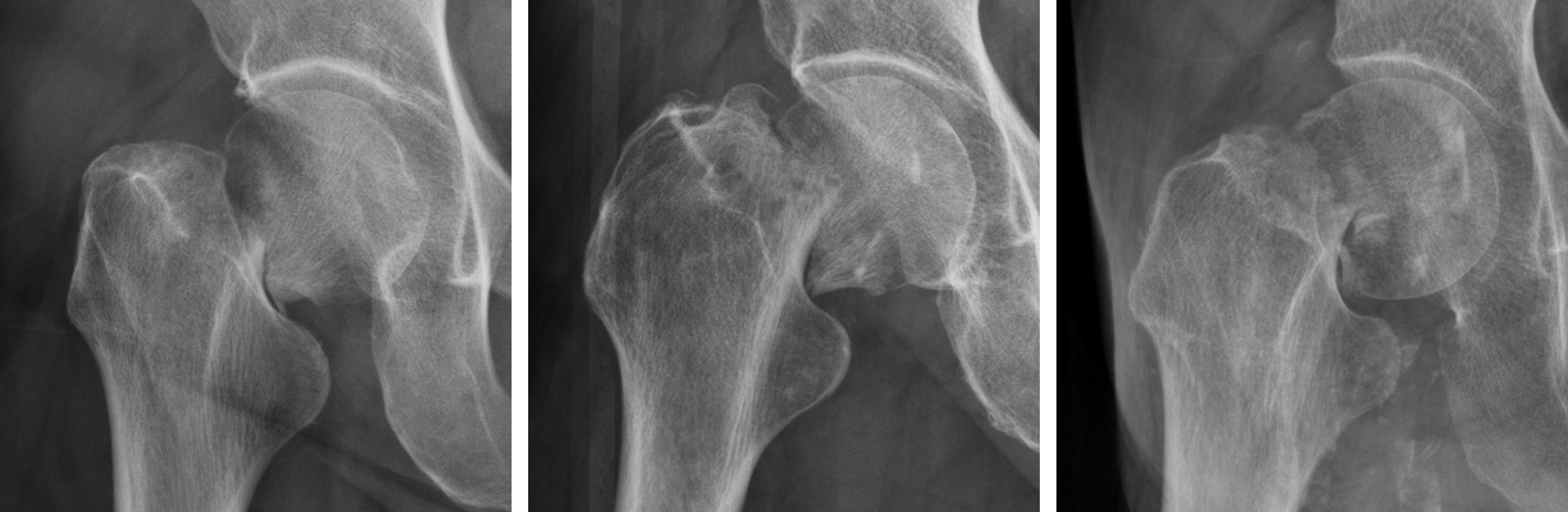}%
\captionsetup{labelfont={color=CLRBlue},font={color=CLRBlue}}
\caption{Examples of crops with detached fractures that have been excluded}%
\label{crop_frac}%
\end{subfigure}\hfill%

\captionsetup{labelfont={color=CLRBlue},font={color=CLRBlue}}
\caption{\wdairv{Examples of valid and invalid crops.}}
\label{crop_examppic}
\end{figure*}

}

\end{document}